\pdfoutput=1
\documentclass[conference]{IEEEtran}
\IEEEoverridecommandlockouts
\usepackage{amsmath,amssymb,amsthm}
\usepackage{graphicx}
\usepackage{booktabs}
\usepackage{xcolor}
\usepackage{cite}
\usepackage[hidelinks]{hyperref}

\newcommand{\svla}{S-VLA}
\newcommand{\lvla}{L-VLA}
\newcommand{\parhead}[1]{\vspace{1mm}\noindent\textbf{#1}}
\newcommand{\firstparhead}[1]{\noindent\textbf{#1}}

\begin{document}

\title{VLA Handover: Failure-Prediction-Based Switching\\from On-Robot to Server VLA Models%
\thanks{This work has been submitted to the IEEE for possible publication. Copyright may be
transferred without notice, after which this version may no longer be accessible. This work is
supported in part by JST-ALCA-Next Japan Grant Number JPMJAN24F1.}}

\author{\IEEEauthorblockN{Takayuki Nishio}
\IEEEauthorblockA{School of Engineering, Institute of Science Tokyo, Tokyo 152-8550, Japan\\
Email: nishio@ict.eng.isct.ac.jp}}

\maketitle

\begin{abstract}
Large vision-language-action (VLA) models generalize well but are difficult to run in real time on a
robot, and small VLAs that run there fail more often. Remote inference over beyond-5G and 6G
networks lets a robot use a large VLA on a server, but querying the server for every action chunk
limits the number of robots per server. We propose VLA handover: the robot runs a small VLA (S-VLA)
and switches to a large VLA (L-VLA) on a server only when the S-VLA is likely to fail. We formulate the switching as minimizing the request cost under a
required success probability and, since this problem cannot be solved directly, propose a heuristic
framework that retrains neither VLA. In simulation with SmolVLA and $\pi_{0.5}$ on three of the four LIBERO
suites, VLA handover raised the success rate by 11--22 percentage points over the S-VLA alone and
sent 49--77\% fewer requests than server-only control at a 2--4 percentage-point lower success rate, and its
completion time grew less with the network latency.
\end{abstract}


\section{Introduction}
Robot foundation models have shown strong
performance in robot manipulation, in particular vision-language-action (VLA)
models~\cite{openvla,pi05}, which map camera images and a natural-language instruction directly
to robot actions. These models rely on vision-language models with billions of parameters and are
therefore difficult to run in real time on the computer mounted on a robot. Small VLAs such as
SmolVLA~\cite{smolvla} are designed to be deployed on consumer-grade GPUs or CPUs, but they
generalize less well: on the LIBERO-Long suite~\cite{libero}, SmolVLA succeeded in 35.0\% of the
episodes in our evaluation, whereas $\pi_{0.5}$~\cite{pi05} succeeded in 93.2\% (Sec.~\ref{sec:eval}).

\begin{figure}[t]
\centering
\includegraphics[width=\columnwidth]{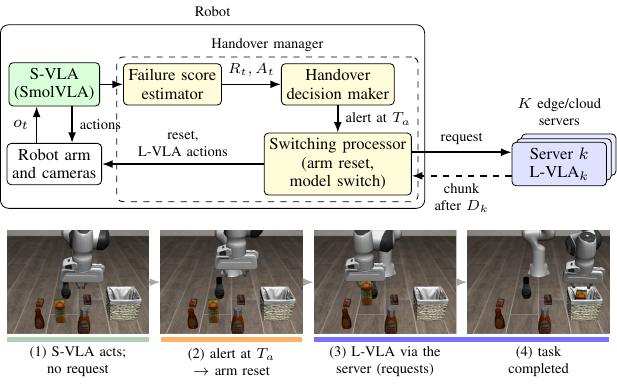}
\caption{VLA handover. Top: system model; the handover manager decides when to hand control from the
\svla\ to an \lvla\ and to which of the $K$ servers ($K=1$ in Sec.~\ref{sec:method}). Bottom: an
episode; the \svla\ sends no request until a failure alert, and the arm is then reset and an \lvla\
takes over.}
\label{fig:system}
\end{figure}

Remote control of robots over low-latency, high-reliability beyond-5G and 6G
networks~\cite{itu2160} allows a robot to use a large VLA: the robot sends the
information required for inference, such as images and its state, to a VLA on an edge or cloud
server, and the server returns the output actions. As the number of robots grows, however, the
communication traffic and the server load become the bottleneck and limit the number of robots that
one server can accommodate. In our setting ($\pi_{0.5}$ on an NVIDIA GB10 without batching), a
server accommodates only about two robots when every action chunk is inferred on the server; the
number depends on the server but is inversely proportional to the requests per robot. In
addition, the communication and computation latency delays each action; in dynamic scenes, or when
the robot cannot hold still while waiting, it can also lower the task success rate.

Existing schemes that combine an on-robot model with a server VLA involve the server in the
inference of every action chunk. RoboECC~\cite{roboecc}, RAPID~\cite{rapid}, and
ComVLA~\cite{comvla} partition the inference of one VLA between the edge and the cloud, adapting the
split point to the network bandwidth or the number of transmitted visual tokens to the channel
capacity. AsyncVLA~\cite{asyncvla} runs a large VLA on a workstation and an on-robot edge adapter
asynchronously, and CloudEdgeVLA~\cite{cloudedgevla} combines delayed cloud features with current
local vision; both train the server and on-robot models jointly. The requests of these schemes do
not decrease when the on-robot model would succeed, and the latter two cannot be applied to existing
VLAs without retraining. Using the server only when it is needed raises the question of when to
switch, which is a trade-off: a switch that is too early sends requests in episodes that the small
VLA would complete alone, whereas a switch that is too late comes after a critical mistake of the
small VLA or too close to the deadline, so that even the large VLA cannot complete the task.

Related decisions have been studied in other settings. In cloud robotics, Chinchali \emph{et
al.}~\cite{chinchali} learn when to offload perception inference from the uncertainty of the
on-robot model and the network cost, and Penmetcha \emph{et al.}~\cite{penmetcha} predict whether to
offload an application from the execution time; these works decide per inference of a perception or
application module, whereas our decision concerns which policy controls the robot for the rest of a
closed-loop episode. In interactive imitation learning, a robot requests a human supervisor when it
is at risk, under a budget of interventions~\cite{thriftydagger}, with fewer
switches~\cite{lazydagger}, or from a limited pool of remote supervisors shared by a
fleet~\cite{fleetdagger}; the helper is a human, and the communication and computation latency of
the help is not modeled.

In this paper, we propose \emph{VLA handover}: the robot runs a small VLA (\svla) and switches to a
large VLA (\lvla) on a server only when the \svla\ is likely to fail.
Fig.~\ref{fig:system} illustrates the system and an example episode. This paper makes the following contributions.

\firstparhead{Formulation and framework.} We formulate VLA handover, i.e., deciding when to switch from an on-robot \svla\ to an \lvla\
and to which of the \lvla s on multiple servers, as the minimization of the expected request cost
subject to a required task success probability (Sec.~\ref{sec:formulation}). Since this problem
cannot be solved directly, we propose a heuristic single-server framework in which a handover manager
estimates failure scores of the \svla, decides from them when to switch, and resets the arm and
switches the model (Sec.~\ref{sec:method}); neither VLA is retrained.

\firstparhead{Proof-of-concept experiments.} In a proof of concept with SmolVLA and $\pi_{0.5}$ on four LIBERO suites under an emulated
latency, we evaluate the requests, the success rate, and the completion time against the alert
threshold and the latency, and report both the suites where VLA handover reduces requests at a small
loss of success and the suite where it falls well short of server-only control (Sec.~\ref{sec:eval}).

\section{System Model and Problem Formulation}\label{sec:formulation}
\subsection{System Model}
\firstparhead{VLA models.} A VLA model takes the robot state, camera images, and a natural-language
instruction as input. A vision-language model (VLM) processes these inputs, and an action head
converts the VLM output into robot actions. With action chunking, the action head generates a chunk
of $H$ actions for the next $H$ control steps from a single observation; the robot executes
$n$ ($\le H$) of them and then queries the model again with a new observation $o_t$. A request to a
remote VLA therefore carries the images, the state, and the instruction, and its response carries a
chunk of actions.

\parhead{System.} Fig.~\ref{fig:system} shows the system, which consists of a robot, an \svla\
$\pi^{\mathrm{S}}$ and a handover manager on the robot, and $K$ servers, each running an \lvla\
$\pi^{\mathrm{L}}_k$, $k\in\mathcal{K}=\{1,\dots,K\}$. Time is indexed by control steps $t$ with period $\Delta$, and an episode
ends at task success or at the step limit $T_{\max}$, which we treat as a task requirement
(Sec.~\ref{sec:eval} discusses its length). Every $n$ steps, i.e., at each chunk boundary, the handover manager
decides from the history $h_t$ of observations and model inferences whether the \svla\ keeps
control ($u_t=0$) or $\pi^{\mathrm{L}}_k$ takes over ($u_t=k\in\mathcal{K}$). After a handover,
every chunk inferred by $\pi^{\mathrm{L}}_k$ is one request to server $k$; each request returns
after the latency $D_k=D_k^{\mathrm{up}}+D_k^{\mathrm{comp}}+D_k^{\mathrm{down}}$, the sum of the
uplink, computation, and downlink latencies, and incurs the cost $c_k$ for communication and
computation. Let $N_k(\phi)$ be the number of requests to server $k$ in an episode under the
switching policy $\phi: h_t\mapsto u_t$, and $N_{\mathrm{req}}=\sum_{k}N_k$ the total number of
requests per episode.

\subsection{Problem Formulation}
The decision variables are when to switch and to which \lvla, both of which are determined by
$\phi$. We minimize the expected request cost, i.e., the cost per request times the number of
requests per episode, subject to a required task success probability $P_{\mathrm{req}}$:
\begin{equation}
\min_{\phi}\ \mathbb{E}\Big[\sum_{k\in\mathcal{K}} c_k\,N_k(\phi)\Big]
\quad\mathrm{s.t.}\quad P_{\mathrm{s}}(\phi)\ge P_{\mathrm{req}},
\label{eq:problem}
\end{equation}
where $P_{\mathrm{s}}(\phi)$ is the task success probability under $\phi$. $P_{\mathrm{req}}$ acts as a
task-level QoS requirement, and $N_{\mathrm{req}}$ determines the load that each robot places on the
network and the server. The choice of the
\lvla\ affects both the objective and the constraint: a more capable \lvla\ recovers more states, whereas a distant or loaded
server has a larger $D_k$ and $c_k$.

We expect $P_{\mathrm{s}}$ to decrease with the time until a handover takes effect,
i.e., the switching time plus $D_k$, in episodes that the \svla\ would fail alone, for two reasons.
First, once the \svla\ has made a critical
mistake, such as dropping an object where the gripper cannot reach it, the \lvla\ cannot recover
the episode; we denote by $\tau_0$ the unobservable time after which the episode cannot be
completed. Second, a late switch leaves too few steps before the deadline $T_{\max}$, as does
a long latency when the robot cannot pause. Latency-reduction methods, such as compressing
the transmitted information~\cite{comvla} or scheduling communication and computation resources,
lower $D_k$ and $c_k$ for a given switching policy; they are complementary to our approach, which decides
when and to which \lvla\ to switch. Since $P_{\mathrm{s}}(\phi)$ has no
closed form and $\tau_0$ is not observable, \eqref{eq:problem} cannot be solved analytically or at
run time, and a search over $\phi$ would require rollouts of both VLAs for every candidate policy.
Section~\ref{sec:method} therefore proposes a heuristic framework.

\section{VLA Handover}\label{sec:method}
In general, the handover decision also selects the server and may decline a handover, e.g., under
heavy server or network load. As a proof of concept, we propose a simple handover manager for a
single server ($K=1$) that always hands over once a failure appears likely, so that the decision
reduces to triggering on a failure alert. As Fig.~\ref{fig:system} shows, it consists of three
modules: a failure score estimator, which computes failure scores of the \svla; a handover decision
maker, which decides from the scores when to hand over; and a switching processor, which executes
the handover. We drop the subscript $k$, so that the objective in \eqref{eq:problem} reduces to
$c\,\mathbb{E}[N_{\mathrm{req}}]$.

\subsection{Failure Score Estimator}
The failure score estimator outputs, at each chunk boundary, scores that rise when the \svla\ is
likely to fail. The framework requires only such scores, not a particular method; the scores used by
runtime failure predictors~\cite{fiper,safe} are one choice. We adopt the two scores of failure
prediction at runtime (FIPER)~\cite{fiper}, which need only successful rollouts of the \svla. The
observation score uses random network distillation (RND): a predictor MLP $f$ is trained on
successful rollouts to reproduce a frozen, randomly initialized target MLP $g$ on the embedding $e_t$
that $\pi^{\mathrm{S}}$ computes from $o_t$, i.e., its VLM hidden states averaged over the input
tokens, and $s^{\mathrm{R}}_t=\|f(e_t)-g(e_t)\|_2$ grows for observations unlike those of successful
rollouts. The action score, the action-chunk entropy (ACE), samples $B$ chunks from the \svla\ for
the same observation, converts each into the end-effector positions $x^{(j)}_{t+i}$, $i=0,\dots,H-1$,
$j=1,\dots,B$, that it commands, and averages over the chunk the entropy of the $B$ positions at each
step, estimated by binning:
\begin{equation*}
s^{\mathrm{A}}_t=\frac{1}{H}\sum_{i=0}^{H-1}\Big(-\sum_{c}q_{c,i}\log_2 q_{c,i}\Big),
\end{equation*}
where $q_{c,i}$ is the fraction of $x^{(1)}_{t+i},\dots,x^{(B)}_{t+i}$ in grid cell $c$; it grows
when the \svla\ is uncertain about what to do. Each score is summed over the last $w_{\mathrm{R}}$ or
$w_{\mathrm{A}}$ inferences (fewer at the start of an episode),
$R_t=\sum_{m=0}^{w_{\mathrm{R}}-1}s^{\mathrm{R}}_{t-mn}$ and
$A_t=\sum_{m=0}^{w_{\mathrm{A}}-1}s^{\mathrm{A}}_{t-mn}$.

\subsection{Handover Decision Maker}
The handover decision maker decides from the scores when to switch. FIPER predicts failures and
does not decide handovers, but its rule for raising a failure alert carries over to this decision,
and we design the decision maker after it. An alert is raised only when both windowed scores exceed
their thresholds, because either score alone also rises in successful rollouts, e.g., in novel but
harmless situations or with multimodal demonstrations~\cite{fiper}, and the handover occurs at the
first alert,
\begin{equation*}
T_a=\min\{t:\ R_t>\kappa\,\theta_{\mathrm{R}}(t),\ A_t>\kappa\,\theta_{\mathrm{A}}(t)\}.
\end{equation*}
The thresholds $\theta_{\mathrm{R}}(t)$ and $\theta_{\mathrm{A}}(t)$ are calibrated as in FIPER on
successful rollouts, as the $(1-\delta)$ quantile of the per-episode maximum of each score, as a
conformal prediction band over time, or as the $(1-\delta)$ quantile at each time; the first two
bound the probability of an alert on a new successful rollout by $\delta$~\cite{fiper}. We use
$\delta=0.1$, the upper end of the range in~\cite{fiper}. To trace the trade-off, we also scale both
thresholds by a common factor $\kappa$ ($\kappa=1$ for the calibrated thresholds); since the two
scores are not proportional, this is only one of many one-parameter families. We label each $\kappa$
by $\alpha$: $\kappa$ is the $(1-\alpha)$ quantile of the per-episode maximum of
$\min\{R_t/\theta_{\mathrm{R}}(t),A_t/\theta_{\mathrm{A}}(t)\}$ over the successful \svla\ rollouts in
the configuration-selection and development (dev) splits (Sec.~\ref{sec:eval}), so that at most a fraction $\alpha$
of them receive an alert.

Unlike the general formulation, in which $u_t$ may change again, the \lvla\ controls the robot for
the rest of the episode after a handover, so that
\begin{equation}
\begin{aligned}
\mathbb{E}[N_{\mathrm{req}}]\approx{}&\big[(1-P_{\mathrm{s}}^{0})\mathrm{TPR}(\kappa)+P_{\mathrm{s}}^{0}\mathrm{FPR}(\kappa)\big]\\
&\times\mathbb{E}\!\left[\tfrac{T_{\mathrm{end}}-T_a-r}{n}\,\middle|\,\text{alert}\right],
\end{aligned}
\label{eq:requests}
\end{equation}
where $P_{\mathrm{s}}^{0}$ is the success probability of the \svla\ alone, i.e., $P_{\mathrm{s}}$ under $u_t\equiv0$, TPR and FPR are the true- and
false-positive rates of the alert, i.e., the fraction of episodes that receive an alert among those
the \svla\ would fail alone and among those it would complete alone, respectively, $T_{\mathrm{end}}$ is the end of the
episode, and $r$ is the number of steps of the arm reset described below; the approximation ignores
rounding. Lowering $\kappa$ raises the TPR and moves $T_a$ earlier, which leaves more time before
$\tau_0$ but also more steps, and thus requests, in handed-over episodes that still fail, and raises
the FPR and thus the requests from episodes that the \svla\ would complete alone, whose success the
handover can also lower (Sec.~\ref{sec:object}). Switching earlier than the alert of the calibrated
thresholds is therefore equivalent to lowering $\kappa$.

Although FIPER was designed for failure prediction, its threshold rule reflects both sides of this
trade-off, which is why we adopt it as the heuristic decision. For the first two threshold types, its calibration bounds the FPR, which
here corresponds to the requests from unnecessary handovers, and FIPER selects the window sizes and the threshold type by the timestep-wise accuracy
$\mathrm{TWA}=\frac{1}{2}[\mathrm{TPR}(1-\mathrm{DT})+1-\mathrm{FPR}]$, where DT is the mean detection
time of the detected failures normalized by the maximum episode length~\cite{fiper}; TWA thus
rewards early alerts, which here leave the \lvla\ more time. TWA does not measure whether the \lvla\ can
still recover at the alert. Since the switching time is $T_a$, the
policy $\phi$ is determined by $\kappa$, and \eqref{eq:problem} becomes a one-dimensional problem:
among the values of $\kappa$ whose success probability is at least $P_{\mathrm{req}}$, find the one
with the fewest expected requests. We do not solve it; we use $\kappa=1$, i.e., the calibrated
thresholds, for the reasons above, and Fig.~\ref{fig:requests} shows the requests and the success rate
over $\kappa$.

\subsection{Switching Processor}
At the switch, the switching processor resets the arm and then switches the model that controls the
robot. (i) Arm reset: it opens the gripper, moves vertically to the initial end-effector height, and
then moves to the initial position and orientation with proportional end-effector commands to the
operational-space controller~\cite{osc}. The reset lasts at most $r_{\max}=80$ steps, which count
toward $T_{\max}$. (ii) Model switch: it sends the current observation to the server, holds the
robot still for $D$, executes the returned chunk, and sends the next request every $n$ executed
steps. A chunk that arrives $d$ executed steps after its request is executed from index $d$; we call
this execution the pipeline. The arm
reset would be unnecessary if the \lvla\ could recover from arbitrary states; we include it in this
proof of concept because, at the same alerts, it raised the success rate of VLA handover by
5.4--15.2 percentage points (pt) and reduced the requests in all four suites (Tables~\ref{tab:main}
and~\ref{tab:suites}).

\begin{figure}[t]
\centering
\includegraphics[width=\columnwidth]{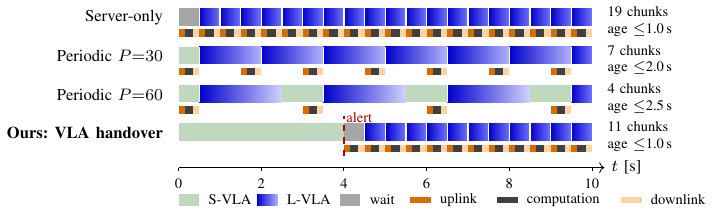}
\caption{Control timelines (conceptual; $D=0.5$\,s for visibility, $H=50$ steps, alert at 4\,s, arm
reset omitted). Thick bars: model in control (lighter: \lvla\ actions computed from older
observations); thin bars: uplink, computation, and downlink of each request.}
\label{fig:timing}
\end{figure}

Fig.~\ref{fig:timing} compares the control timeline of VLA handover with those of the baselines in
Sec.~\ref{sec:eval}. VLA handover sends no request until the alert and then requests every chunk;
each executed action is computed from an observation at most $D+n\Delta$ old.

\section{Evaluation}\label{sec:eval}

\subsection{Setup}
\firstparhead{Benchmark and policies.} LIBERO~\cite{libero} is a simulated manipulation benchmark with four
suites of 10 tasks each: LIBERO-Spatial, -Object, and -Goal vary the spatial layout, the object, and
the task goal, respectively, and LIBERO-Long (also called LIBERO-10) consists of long-horizon tasks. The \svla\ is \nolinkurl{HuggingFaceVLA/smolvla_libero} (all 32 layers of
SmolVLM2-500M, unlike the 16-layer configuration in~\cite{smolvla}) and the \lvla\ is
\nolinkurl{lerobot/pi05_libero_finetuned}, both public checkpoints fine-tuned on LIBERO and used
without retraining in LeRobot~\cite{lerobot} with $H=50$, $n=10$, and $\Delta=0.05$\,s. On an NVIDIA
GB10, $\pi_{0.5}$ serves about 4.4 inferences/s (229\,ms per inference, batch 1), and the \svla\
with FIPER ($B=16$, ACE on end-effector positions) takes 0.371\,s per chunk, within the 0.5\,s
execution time of a chunk, so we neglect it in the latency. For each suite, we chose the window sizes and
the threshold type that maximized TWA on the configuration-selection split: a constant threshold
(the quantile of the per-episode maximum) on LIBERO-Object and -Goal, the per-time quantile on
LIBERO-Spatial, and the conformal band on LIBERO-Long. The thresholds were calibrated with $\delta=0.1$; on LIBERO-Object,
$\theta_{\mathrm{R}}$ and $\theta_{\mathrm{A}}$ are 0.360 and 17.74. The successes that define $\alpha$ were run to the LIBERO default limit in the
configuration-selection split and to $T_{\max}$ in the dev split (on LIBERO-Long, to 520 steps in
both, with half of the dev split).

\parhead{Latency and requests.} A request carries two $360\times360$ RGB images (about 0.78\,MB
uncompressed), the 8-dimensional state, and the instruction, and a response carries $50\times7$
actions. We emulate the total latency $D$, i.e., uplink, computation, and downlink together, instead
of deriving it from the payload and the link rate. Mixtures of log-normals describe measured
latencies more closely~\cite{volos}; for simplicity, we draw $D$ per request from a single
log-normal distribution with median $\bar D=0.25$\,s and a coefficient of variation (CV) of 0.3, with
common random numbers across schemes and independently of the server load. Values of $D$ below the
GB10 inference time (229\,ms) assume a faster server. The scene is quasi-static: the simulator is paused
while the robot waits, so waiting counts in the completion time but not in $T_{\max}$.

\parhead{Baselines.} \emph{Device-only} and \emph{server-only} control mark the two ends of the
request count; server-only control waits for the first response and then uses the pipeline of
Sec.~\ref{sec:method}. \emph{Periodic querying} uses the server without an alert, as the existing
collaboration schemes do: it sends the current observation every $P$ steps while the \svla\ keeps
acting, executes each response from index $d$ until it is exhausted or replaced, and then resumes
the \svla. While the next response arrives before the current chunk is exhausted
($P+D/\Delta<H$, i.e., $P\lesssim45$), the \lvla\ controls almost all steps, so periodic querying with
$P$ up to about 50 is essentially server-only control that executes up to $P$ actions of each chunk,
computed from observations up to $P\Delta+D$ old. $P\in\{10,20,30,50,100\}$ is selected on the
dev split as the period with the fewest requests within 5\,pt of server-only control.
Two ablations change one module of VLA handover:
\emph{w/o arm reset} hands over at the same alerts without the reset, and \emph{fixed step}
replaces the alert by a fixed handover step $t_h$, the mean of the completion steps of the
successful device-only dev episodes rounded up to a multiple of $n$ (150, 100, 110, and 300 on
LIBERO-Object, -Goal, -Spatial, and -Long) and, on LIBERO-Object, also their maximum (460).

\parhead{Metrics and protocol.} The primary metrics are the success rate (95\% Wilson
interval) and the number of requests per episode, i.e., the objective in
\eqref{eq:problem} with a constant cost. We also
report the server fraction, i.e., the fraction of steps controlled by the \lvla, and the completion
time, i.e., the mean wall-clock time of successful episodes including waits. Initial placements are
sampled from disjoint seeds for FIPER calibration, configuration selection, dev, and evaluation. Handover
schemes resume from the saved state at the alert of the device-only run, and each episode is run
once. The step limits are longer than the LIBERO defaults (280, 300, 220, and 520 steps on LIBERO-Object,
-Goal, -Spatial, and -Long), which do not assume a switch or a
resumption within an episode: on LIBERO-Object, successful handed-over episodes took a median of 306
steps, and at the default limit VLA handover (82.4\%) was close to device-only control (82.0\%).

\subsection{Main Results on LIBERO-Object}\label{sec:object}
\begin{table}[t]
\centering
\caption{LIBERO-Object evaluation split (500 episodes), $T_{\max}=500$,
$\bar D=0.25$\,s. Success in \% with 95\% Wilson interval}
\label{tab:main}
\footnotesize
\setlength{\tabcolsep}{3pt}
\begin{tabular}{lcccc}
\toprule
Scheme & Success & Req./ep. & Srv.\ frac. & Compl.\,[s] \\
\midrule
Device-only & 83.8 [80.3, 86.8] & 0 & 0 & 7.36 \\
Server-only & 96.8 [94.9, 98.0] & 14.99 & 1.00 & 6.96 \\
Periodic $P=10$ & 96.4 [94.4, 97.7] & 15.00 & 0.96 & 6.62 \\
Periodic $P=20$ & 91.8 [89.1, 93.9] & 8.97 & 0.97 & 7.03 \\
Periodic $P=30$ & 86.8 [83.6, 89.5] & 6.59 & 0.97 & 6.82 \\
\textbf{Ours: VLA handover} & \textbf{94.4 [92.0, 96.1]} & \textbf{3.40} & \textbf{0.18} & \textbf{8.51} \\
\quad -- w/o arm reset & 89.0 [86.0, 91.5] & 4.92 & 0.26 & 7.64 \\
\quad -- fixed step $t_h{=}150$ & 91.6 [88.8, 93.7] & 5.17 & 0.25 & 8.90 \\
\quad -- fixed step $t_h{=}460$ & 83.6 [80.1, 86.6] & 0.01 & 0.00 & 7.32 \\
\bottomrule
\end{tabular}\\[2pt]
\parbox{\columnwidth}{\scriptsize Srv.\ frac.: fraction
of steps controlled by the \lvla. Compl.: mean completion time of successful episodes.}
\end{table}

Table~\ref{tab:main} shows how VLA handover reduces the server requests while keeping the success
rate close to that of server-only control. VLA handover succeeded in 94.4\%, 10.6\,pt above
device-only control and 2.4\,pt below server-only control, with 3.40 requests per episode, 77.3\%
fewer than server-only control. Dividing the throughput of $\pi_{0.5}$ (4.4 inferences/s) by the mean
request rate of a robot, i.e., its requests per second of wall-clock time, gives about 12 robots per
server for VLA handover and 2 for server-only control, assuming full utilization and no queueing;
since the requests concentrate in handed-over episodes, the instantaneous load can exceed this
average. The price is a longer completion time (8.51 vs.\ 6.96\,s). VLA handover thus came within a
few points of server-only control with about a quarter of its requests, which lets a server
accommodate several times more robots.

Fixed-step handover with $t_h=150$ succeeded in 91.6\% with 5.17 requests: it handed over 97 of the
419 episodes that device-only control completed and lost 25 of them, whereas VLA handover handed over
28 and lost none, although VLA handover recovered fewer of the device-only failures that FIPER alerted after
step 150 or missed (13 vs.\ 24 of 29; 40 each of the other 52). With $t_h=460$, the handover came too
late for the \lvla. The alert thus helped mainly by not handing over episodes that the \svla\ would
complete.

\begin{figure}[t]
\centering
\includegraphics[width=\columnwidth]{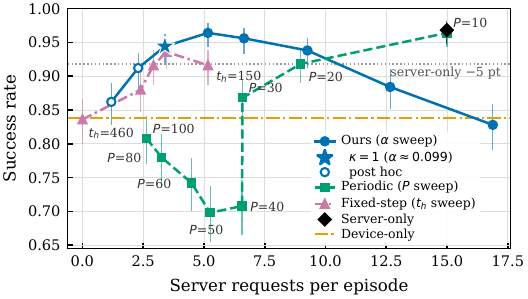}
\caption{Requests per episode vs.\ success rate (LIBERO-Object evaluation split, $T_{\max}=500$,
$\bar D=0.25$\,s). VLA handover: $\alpha$ from 0.02 (left) to 0.9 (right); star:
$\kappa=1$. Fixed-step: $t_h$ from 460 (left) to 150 (right). $\alpha=0.02$ and 0.05, $P=40$, 60,
and 80, and $t_h=200$, 250, and 300 were added after the evaluation split had been examined. Bars:
95\% Wilson intervals.}
\label{fig:requests}
\end{figure}

\parhead{Trade-off over the threshold.} Fig.~\ref{fig:requests} shows that, between 2 and 7 requests
per episode, the curve of VLA handover lies above that of periodic querying. The fixed-step curve
lies below it, but $t_h=200$ came close (93.6\% [91.1, 95.4] with 3.46 requests); on LIBERO-Object,
the gain of the alert over the best fixed step thus lies within the intervals. Episodes handed over on a false alert succeeded in
28/28 at $\alpha\approx0.099$ but in 307/382 at $\alpha=0.9$, where the success rate fell to 82.8\%
with 16.88 requests. At $\alpha=0.2$, VLA handover matched periodic querying with $P=10$ (96.4\%) with
5.16 requests, a point identified after examining the evaluation split. The periodic curve is not
monotonic (86.8, 69.8, and 80.8\% at $P=30$, 50, and 100), possibly because of stale actions around
$P=40$--50 and more \svla\ control at larger $P$; we did not isolate the cause.

\subsection{Other LIBERO Suites}
\begin{table}[t]
\centering
\caption{Other suites (evaluation splits). Goal and Spatial: 300 episodes each; Long: 500. $\kappa=1$; $P$ selected on dev. Success in \% at $T_{\max}$}
\label{tab:suites}
\footnotesize
\setlength{\tabcolsep}{2pt}
\begin{tabular}{llccc}
\toprule
Suite & Scheme & Success & Req./ep. & Srv.\ frac. \\
\midrule
Goal & Device-only & 76.3 & 0 & 0 \\
$T_{\max}{=}450$ & Server-only & 94.0 & 13.15 & 1.00 \\
 & Periodic $P{=}30$ & 91.0 & 5.01 & 0.96 \\
 & \textbf{Ours: VLA handover} & \textbf{91.0} & \textbf{5.32} & \textbf{0.29} \\
 & \quad -- w/o arm reset & 77.7 & 7.66 & 0.41 \\
 & \quad -- fixed step $t_h{=}100$ & 87.0 & 7.47 & 0.37 \\
\midrule
Spatial & Device-only & 67.3 & 0 & 0 \\
$T_{\max}{=}400$ & Server-only & 94.0 & 12.55 & 1.00 \\
 & Periodic $P{=}10$ & 94.7 & 12.28 & 0.95 \\
 & \textbf{Ours: VLA handover} & \textbf{89.7} & \textbf{6.46} & \textbf{0.36} \\
 & \quad -- w/o arm reset & 80.0 & 9.04 & 0.52 \\
 & \quad -- fixed step $t_h{=}110$ & 80.7 & 5.92 & 0.30 \\
\midrule
Long & Device-only & 35.0 & 0 & 0 \\
$T_{\max}{=}700$ & Server-only & 93.2 & 29.17 & 1.00 \\
 & Periodic $P{=}10$ & 92.0 & 29.41 & 0.98 \\
 & \textbf{Ours: VLA handover} & \textbf{63.6} & \textbf{19.17} & \textbf{0.38} \\
 & \quad -- w/o arm reset & 48.4 & 25.80 & 0.49 \\
 & \quad -- fixed step $t_h{=}300$ & 64.2 & 19.64 & 0.37 \\
\bottomrule
\end{tabular}
\end{table}

Table~\ref{tab:suites} shows that VLA handover worked on LIBERO-Goal and -Spatial as on LIBERO-Object:
it raised the success rate over device-only control by 14.7 and 22.4\,pt, to 3.0 and 4.3\,pt below
server-only control, with 60\% and 49\% fewer requests, and fixed-step handover was 4.0 and 9.0\,pt
below it. Periodic querying reached the same success with slightly fewer requests on LIBERO-Goal
($P=30$) but needed about twice the requests for a 5\,pt higher success rate on LIBERO-Spatial
($P=10$). At $\kappa=1$, $\alpha$ was 0.099 on LIBERO-Goal and 0.248 on LIBERO-Spatial, whose
per-time quantile has no bound on the alert rate~\cite{fiper}. On LIBERO-Long, however, VLA handover reached only 63.6\% ([59.3, 67.7]) against 93.2\% for server-only
control and 92.0\% for periodic querying. Since the \svla\ succeeded
in only 35\%, 71\% of the episodes were handed over, and the \lvla\ completed only 49\% of them,
against 74--80\% on the other suites (Table~\ref{tab:strata}). The handed-over episodes that still
failed also dominate the requests, as Sec.~\ref{sec:strata} shows.

\subsection{Where the Requests Come From}\label{sec:strata}
\begin{table}[t]
\centering
\caption{Requests of VLA handover by handover (H) and outcome. Last column: requests per episode,
all and without handed-over failures}
\label{tab:strata}
\footnotesize
\setlength{\tabcolsep}{2.2pt}
\begin{tabular}{lccccc}
\toprule
 & \multicolumn{2}{c}{H, success} & \multicolumn{2}{c}{H, failure} & Req./ep. \\
Suite & Ep. & Req./ep. & Ep. & Req./ep.\ (share) & all $\to$ w/o H-fail. \\
\midrule
Object & 16.2\% & 14.2 & 4.6\% & 23.7 (32\%) & 3.40 $\to$ 2.31 \\
Goal & 25.0\% & 11.9 & 9.0\% & 26.0 (44\%) & 5.32 $\to$ 2.98 \\
Spatial & 41.7\% & 9.5 & 10.3\% & 24.4 (39\%) & 6.46 $\to$ 3.94 \\
Long & 34.6\% & 20.7 & 36.2\% & 33.2 (63\%) & 19.17 $\to$ 7.14 \\
\bottomrule
\end{tabular}\\[2pt]
\parbox{\columnwidth}{\scriptsize Episodes without a handover send no request (succeeded: 78.2,
66.0, 48.0, 29.0\%; failed: 1.0, 0, 0, 0.2\%).}
\end{table}

Table~\ref{tab:strata} shows that handed-over failures, which keep the \lvla\ running until
$T_{\max}$, were 4.6--36.2\% of the episodes but produced 32--63\% of the requests; on LIBERO-Long,
they made VLA handover send 0.66 of the requests of server-only control (0.23--0.51 on the
others), consistent with \eqref{eq:requests}. Stopping a handover once recovery
becomes unlikely would reduce the requests to the last column of Table~\ref{tab:strata} without
lowering the success rate; this is a counterfactual on the recorded episodes and requires predicting
at run time whether the \lvla\ can still complete the task, which we leave for future work.

\subsection{Effect of Latency}\label{sec:latency}
\begin{figure}[t]
\centering
\includegraphics[width=\columnwidth]{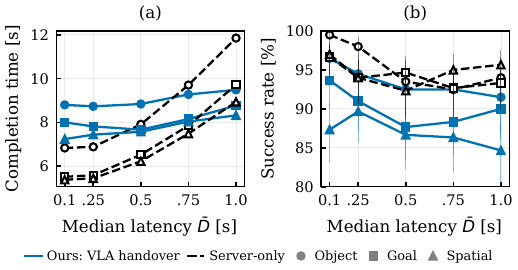}
\caption{Effect of the median latency $\bar D$ (CV 0.3) on (a) the mean completion time of successful
episodes and (b) the success rate (95\% Wilson intervals). LIBERO-Object 200 and -Goal and -Spatial
300 episodes each, the same for both schemes.}
\label{fig:latency}
\end{figure}

We varied $\bar D$ over $\{0.1, 0.25, 0.5, 0.75, 1.0\}$\,s on LIBERO-Object, -Goal, and -Spatial on
the episodes run with server-only control; the robot holds still once it has executed $n$ steps
after a request whose response has not arrived, which differs from the pipeline only for
$D>n\Delta$. Fig.~\ref{fig:latency}a shows that, from 0.1 to 1.0\,s, the mean completion time grew
by 3.6--5.0\,s under server-only control, since every chunk incurs $D$, but by less than 1.2\,s
under VLA handover, since episodes without an alert run on the \svla\ and incur no latency
(2.7--5.8\,s in handed-over episodes alone); VLA handover, whose handovers add the arm reset and server
control, was slower up to 0.5--0.75\,s and faster beyond. Fig.~\ref{fig:latency}b shows that the
success rate of VLA handover was 1.0--5.0\,pt below its value at 0.25\,s for $\bar D\ge0.5$\,s, and
on LIBERO-Spatial it stayed 4.3--11.0\,pt below server-only control. Since episodes without an alert
do not depend on $D$, this drop came only from handed-over episodes; as waits do not count toward
$T_{\max}$, it is presumably due to actions computed from older observations (for $D>n\Delta$, each
chunk is executed from its $(n{+}1)$-th action), which also lowered the success rate of server-only
control by 1.3--5.5\,pt from 0.1 to 1.0\,s. By running most episodes on the \svla, VLA handover
thus keeps its completion time nearly insensitive to the latency, at a small cost in success.

\section{Conclusion}
We proposed VLA handover, which switches from an on-robot \svla\ to a server \lvla\ only when the
\svla\ is likely to fail, together with a request-cost formulation and a heuristic framework that
retrains neither VLA. On three LIBERO suites, it recovered most of the success gap between the \svla\
alone and server-only control with a quarter to a half of the requests, whereas it fell short on
LIBERO-Long, where the \svla\ rarely succeeds. Handed-over failures produced a third to two thirds of
the requests, so stopping a handover once recovery becomes unlikely is the key next step. The
evaluation was limited to simulation with one \lvla, single runs, and a load-independent,
light-tailed latency in quasi-static scenes; server selection, load-aware handover, latency-aware
thresholds, dynamic scenes, and real robots are future work.

\bibliographystyle{IEEEtran}
\bibliography{refs}

\end{document}